\documentclass[11pt,a4paper]{article}
\usepackage[hyperref]{eacl2021}
\usepackage{times}
\usepackage{latexsym}

\usepackage{textcomp}

\usepackage{graphicx}
\usepackage{tabularx}

\usepackage{amsmath}
\DeclareMathOperator*{\argmax}{arg\,max}

\usepackage{subfig}
\newcounter{notecounter}
\newcommand{\enotesoff}{\long\gdef\enote##1##2{}}
\newcommand{\enoteson}{\long\gdef\enote##1##2{{
			\stepcounter{notecounter}
			\large\bf
			\hspace{100cm}\arabic{notecounter} $<<<$ ##1: ##2
			$>>>$\hspace{1cm}}}}
\enoteson
\enotesoff

\usepackage{pifont}

\usepackage{microtype}

\aclfinalcopy 

\title{Addressing Zero-Resource Domains Using Document-Level Context in Neural Machine Translation}

\author{Dario Stojanovski \and Alexander Fraser\\
	Center for Information and Language Processing \\
	LMU Munich, Germany \\
	{\tt \{stojanovski,fraser\}@cis.lmu.de}
}

\date{}

\begin{document}
\maketitle

\begin{abstract}
Achieving satisfying performance in machine translation on domains for which there is no training data is challenging. Traditional 
supervised
domain adaptation is not suitable for addressing such zero-resource domains because it relies on in-domain parallel data. We show that when in-domain parallel data is not available, access to document-level context enables better capturing of domain generalities compared to only having access to a single sentence. Having access to more information provides a more reliable domain estimation.  
We present two document-level Transformer models which are capable of using large context sizes and we compare these models against strong Transformer baselines. We obtain improvements for the two zero-resource domains we study. 
We additionally provide an analysis where we vary the amount of context and look at the case where in-domain data is available.
\end{abstract}

\section{Introduction}

Training robust neural machine translation models for a wide variety of domains is an active field of work. NMT requires large bilingual resources which are not available for many domains and languages. When there is no data available for a given domain, e.g., in the case of web-based MT tools, this is a significant challenge. Despite the fact that these tools are usually trained on large scale datasets, they are often used to translate documents from a domain which was not seen during training. We call this scenario zero-resource domain adaptation and present an approach using document-level context to address it.

When an NMT model receives a test sentence from a zero-resource domain, it can be matched to similar domains in the training data. This is to some extent done implicitly by standard NMT. Alternatively, this matching can be facilitated by a domain adaptation technique such as using special domain tokens and features \cite{kobus2017control,tars2018multi}. However, it is not always easy to determine the domain of a sentence without larger context. Access to document-level context makes it more probable that domain signals can be observed, i.e., words representative of a domain are more likely to be encountered.
We hypothesize that this facilitates better matching of unseen domains to domains seen during training
and provide experimental evidence supporting this hypothesis.

Recent work has shown that contextual information improves MT \cite{miculicich2018documentnmt,voita2019good,maruf2019selective}, often by improving anaphoric pronoun translation quality, which can be addressed well with limited context. However, in order to address discourse phenomena such as coherence and cohesion, access to larger context is preferable. \newcite{voita2019good,voita2019context} were the first to show large improvements on lexical cohesion in a controlled setting using challenge sets. However, previous work did not make clear whether previous models can help with disambiguation of polysemous words where the sense  is domain-dependent.

In this work, we study the usefulness of document-level context for zero-resource domain adaptation (which we think has not been studied in this way before). We propose two novel Transformer models which can efficiently handle large context and test their ability to model multiple domains at once. We show that document-level models trained on multi-domain datasets provide improvements on zero-resource domains. We evaluate on English$\rightarrow$German translation using TED and PatTR (patent descriptions) as zero-resource domains. In addition to measuring translation quality, we conduct a manual evaluation targeted at word disambiguation. We also present additional experiments on
classical domain adaptation where access to in-domain TED and PatTR data is allowed.

Our first proposed model, which we call the domain embedding model (DomEmb) applies average or max pooling over all context embeddings and adds this representation to each source token-level embedding in the Transformer. The second model is conceptually similar to previous work on context-aware NMT \cite{voita2018anaphora,stojanovski2018oracles,miculicich2018documentnmt,zhang2018improving} and introduces additional multi-head attention components in the encoder and decoder in order to handle the context. However, in order to facilitate larger context sizes, it creates a compressed context representation by applying average or max pooling with a fixed window and stride size. We compare our proposed models against previous context-aware NMT architectures and techniques for handling multi-domain setups, and show they improve upon strong baselines. The proposed models encode context in a coarse-grained way. They only have a limited ability to model discourse phenomena such as coreference resolution, so the gains we see in a multi-domain setup show that they encode domain information. Evaluating on multiple and zero-resource domains allows us to show that context can be used to capture domain information.

The contributions of our work can be summarized as follows: we (i) propose two NMT models which are able to handle large context sizes, (ii) show that document-level context in a multi-domain experimental setup is beneficial for handling zero-resource domains, (iii) show the effect of different context sizes and (iv) study traditional domain adaptation with access to in-domain data.

\begin{figure*}
	\centering
	\includegraphics[width=13cm,keepaspectratio]{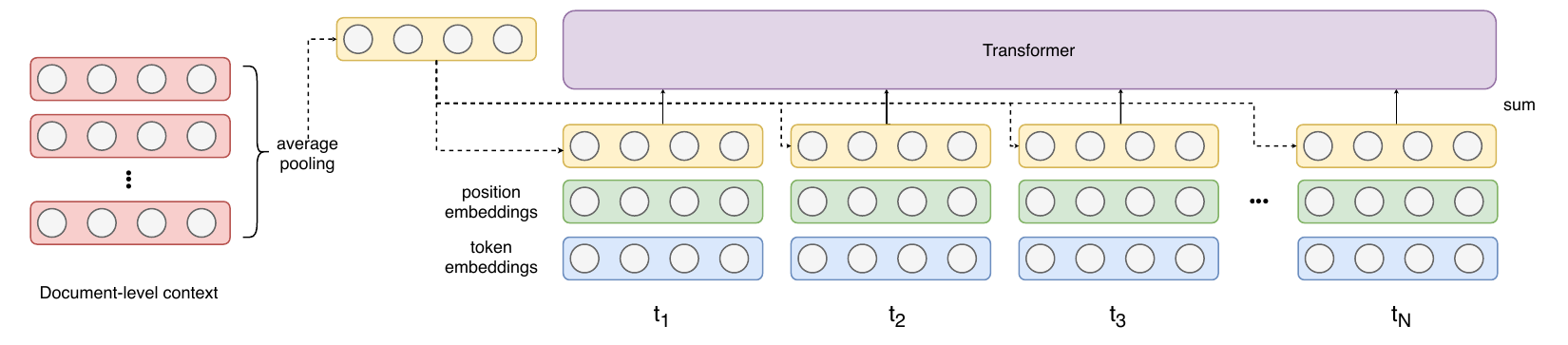}
	\caption{Domain embedding Transformer.}
	\label{fig:domain-embedding-model}
\end{figure*}

\begin{figure}[!ht]
	\centering
	\includegraphics[width=5.5cm,keepaspectratio]{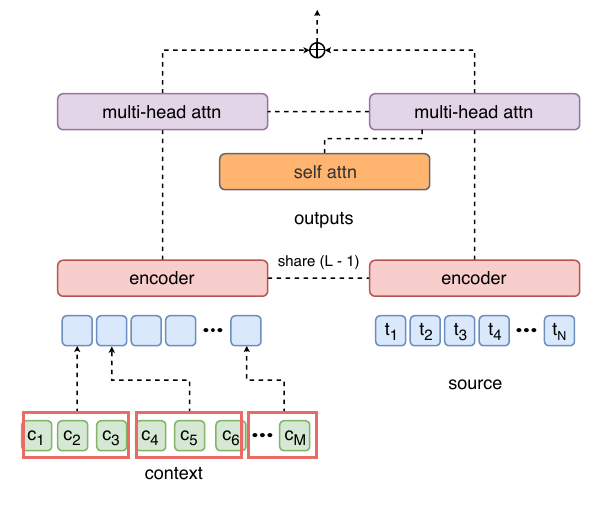}
	\caption{Context-aware Transformer with pooling.}
	\label{fig:ctx-pooling-model}
\end{figure}

\section{Related Work}

\textbf{Domain adaptation}
Several previous works address the problem that standard NMT may fail to adequately model all domains in a multi-domain setup even when all of the domains are known in advance. \newcite{kobus2017control} introduce using domain tags for this problem, a similar method to the domain embedding model in our paper. These domain tags are mapped to corresponding embeddings and are either inserted at the beginning of the sentence or concatenated to the token-level embeddings. The domain embeddings are reserved for specific domains and are fixed for all sentences in a given domain. The number of distinct domain embeddings is limited to the number of known domains. \newcite{tars2018multi} define a similar approach which uses oracle domain tags and tags obtained using supervised methods and unsupervised clustering. However, clustering limits how many domains can be taken into consideration. Furthermore, this approach assumes that sufficient domain information can be obtained from a single sentence alone. Document-level classifiers \cite{xu2007domain} address this problem, but they are not jointly trained with the MT model. Further work in multi-domain MT is \newcite{foster2007mixture} who propose mixture models to dynamically adapt to the target domain, \newcite{foster2010discriminative} who build on this work and include instance weighting,  \newcite{zeng2018multi} where domain-specific and domain-shared annotations from adversarial domain classifiers are used and \newcite{britz2017effective} where a discriminator is used to backpropagate domain signals.

Continued training is an established technique for domain adaptation if access to in-domain resources is possible. The method entails initially training on out-of-domain data, and then continuing training on in-domain data \cite{luong2015stanford}. \newcite{chen2017cost} and \newcite{zhang2018sentence} improve upon this paradigm by integrating a domain classifier or a domain similarity metric into NMT and modifying the training cost based on weights indicating in-domain or out-of-domain data. \newcite{sajjad2017multidomain} and \newcite{farajian2017multidomain} use continued training in a multi-domain setup and propose various ways of fine-tuning to in-domain data. Standard continued training \cite{luong2015stanford} leads to catastrophic forgetting, evident by the degrading performance on the out-of-domain dataset. \newcite{freitag2016fast} address this issue by ensembling the original and the fine-tuned model. We show that our model obtains significant improvements compared to a baseline with the ensembling paradigm. In contrast to these previous works, we do not know the domains during training. Our proposed approaches model the domain implicitly by looking at document-level context. Moreover, we evaluate performance on domains not seen during training.

\newcite{naradowsky2020machine} adapt to unseen domains using bandit learning techniques. The method relies on explicit user feedback which is not always easily available. \newcite{bapna-firat2019nonparametric} propose a retrieval-based method that, at inference time, adapts to domains not seen during training. However, they assume access to in-domain parallel data at inference time, and they retrieve parallel phrases from this in-domain data. In our zero-resource experiments, we have no access to in-domain parallel data.

\textbf{Context-aware NMT} A separate field of inquiry is context-aware NMT which proposes integrating cross-sentence context \cite{tiedemann2017neural,bawden2017evaluating,voita2018anaphora,zhang2018improving,stojanovski2018oracles,miculicich2018documentnmt,tu2018learning,maruf2018document,voita2019good,maruf2019selective,yang2019enhancing,voita2019context,tan2019hierarchical}. These works show that context helps with discourse phenomena such as anaphoric pronouns, deixis and lexical cohesion. \newcite{kim2019when} show that using context can improve topic-aware lexical choice, but in a single-domain setup.

Previous work on context-aware NMT has mostly worked with limited context. \newcite{miculicich2018documentnmt} address the problem by reusing previously computed encoder representations, but report no BLEU improvements by using context larger than 3 sentences. \newcite{zhang2018improving} find 2 sentences of context to work the best. \newcite{maruf2018document} use a fixed pretrained RNN encoder for context sentences and only train the document-level RNN. \newcite{junczys-dowmunt-2019microsoft} concatenates sentences into very large inputs and outputs as in \newcite{tiedemann2017neural}. \newcite{maruf2019selective} propose a scalable context-aware model by using sparsemax which can ignore certain words and hierarchical attention which first computes sentence-level attention scores and subsequently word-level scores. However, for domain adaptation, the full encoder representation is too granular and not the most efficient way to obtain domain signals, for which we present evidence in our experiments. \newcite{stojanovski2019combining,mace2019using} propose a similar approach to our domain embedding model, but they do not investigate it from a domain adaptation perspective. 

To our knowledge, our work is the first at the intersection of domain adaptation and context-aware NMT and shows that document-level context can be used to address zero-resource domains.

\section{Model}

The models we propose in this work are extensions of the Transformer \cite{vaswani2017attention}. The first approach introduces separate domain embeddings applied to each token-level embedding. The second is conceptually based on previous context-aware models \cite{voita2018anaphora,stojanovski2018oracles,miculicich2018documentnmt,zhang2018improving}. Both models are capable of handling document-level context. We modify the training data so that all sentences have access to the previous sentences within the corresponding source document. Access to the document-level context is available at test time as well. Sentences are separated with a special $<$SEP$>$ token from the next sentence. We train and evaluate our models with a 10 sentence context.

\subsection{Domain Embedding Transformer}

The first model is shown in Figure \ref{fig:domain-embedding-model}. It is inspired by \newcite{kobus2017control} which concatenates a special domain tag to each token-level embedding. \newcite{kobus2017control} assume access to oracle domain tags during training. However, at inference, perfect domain knowledge is not possible. Consequently, the domain has to be predicted in advance which creates a mismatch between training and inference. An additional problem is inaccurately predicted domain tags at test time. We modify this approach by replacing the predefined special domain tag with one inferred from the document context. A disadvantage of this approach as opposed to \newcite{kobus2017control} is that there is no clear domain indicator. However, the model is trained jointly with the component inferring the domain which increases the capacity of the model to match a sentence from an unseen domain to a domain seen during training.

The main challenge is producing the domain embedding from the context. We use maximum (DomEmb(max)) or average pooling (DomEmb(avg)) over all token-level context embeddings, both resulting in a single embedding representation. We do not apply self-attention over the context in this model. The intuition is that the embeddings will contain domain information in certain regions of the representation and that this can be extracted by max or average pooling. More domain-specific words will presumably increase the related domain signal. In contrast to a sentence-level model, large context can help to more robustly estimate the domain. Based on preliminary experimental results, we add a feed-forward neural network after the pooled embedding representation in DomEmb(avg), but not in DomEmb(max).
We represent each token as a sum of positional, token-level embeddings and the inferred domain embedding. As the model only averages embeddings, the computational overhead is small. A computational efficiency  analysis is provided in the appendix.

\subsection{Context-Aware Transformer with Pooling}

The second approach (CtxPool) is similar to previous work on context-aware NMT (e.g., \cite{stojanovski2018oracles,zhang2018improving}). The model is outlined in Figure \ref{fig:ctx-pooling-model}. It first creates a compact representation of the context by applying max or average pooling over the context with certain window and stride sizes. The intuition is similar to DomEmb, but pooling over a window provides a more granular representation. We use the concatenation of all context sentences (separated by $<$SEP$>$) as input to CtxPool.

The output of applying max or average pooling over time is used as a context representation which is input to a Transformer encoder. We share the first $L-1$ encoder layers between the main sentence and the context. $L$ is the number of encoder layers. In the decoder, we add an additional multi-head attention (MHA) over the context. This attention is conditioned on the MHA representation from the main sentence encoder. Subsequently, these two representations are merged using a gated sum. The gate controls information flow from the context.

In contrast to DomEmb, CtxPool can be used to handle other discourse phenomena such as anaphora resolution. In this work, we use a window size of 10, suitable for domain adaptation. For anaphora, summarizing ten neighboring words makes it difficult to extract antecedent relationships. Careful tuning of these parameters in future work may allow modeling both local and global context. 

\section{Experiments}

\subsection{Experimental Setup}

We train En$\rightarrow$De models on Europarl, NewsCommentary, OpenSubtitles, Rapid and Ubuntu. TED and PatTR are considered to be zero-resource domains for which we have no parallel data. In additional experiments, we also consider classical domain adaptation where we do use TED and PatTR parallel data in a continued training setup. The models are implemented in Sockeye \cite{Sockeye:17}. The code and the datasets are publicly available.\footnote{\url{https://www.cis.uni-muenchen.de/~dario/projects/zero_domain}} The preprocessing details and model hyperparameters are provided in the appendix.

\subsection{Datasets}

The datasets for some domains are very large. For example, OpenSubtitles contains 22M sentences and PatTR 12M. 
Due to limited computational resources, we randomly sample documents from these domains, ending up with approximately 10\% of the initial dataset size. We keep the original size for the remaining datasets. Dataset sizes for all domains are presented in Table \ref{table:dataset-sizes}. The development and test sets are also randomly sampled from the original datasets. We sample entire documents rather than specific sentences. For TED we use tst2012 as dev and tst2013 as test set. The TED and PatTR dev sets are only used in the fine-tuning experiments where we assume access to in-domain data and are not used in any other experiment.

\begin{table}[ht!]
	\normalsize
	\centering
	\def\arraystretch{1}
	\setlength\tabcolsep{0.6em}
	\begin{tabular}{ l | c | c | c}
		domain         & train& dev  &  test  \\ \hline
		Europarl       &  1.8M    &  3.2K  &  3.0K  \\
		NewsCommentary &  0.3M    &  1.5K  &  1.5K    \\
		OpenSubtitles  &  2.2M    &  2.7K  &  3.3K  \\
		Rapid          &  1.5M    &  2.5K  &  2.5K      \\
		Ubuntu 		   &  11K     &  1.1K  &  0.6K  \\
		\hline
		TED 		   &  0.2M    &  1.7K  &  1.0K    \\
		PatTR	 	   &  1.2M    &  2.0K  &  2.2K  \\
	\end{tabular}
	\caption{Domain datasets sizes in sentences.}
	\label{table:dataset-sizes}
\end{table}

Europarl, NewsCommentary, OpenSubtitles, Rapid and TED are provided with document boundaries. Ubuntu lacks a clear discourse structure and PatTR is sentence-aligned, but provides document IDs. Previous work has shown that context-aware NMT performance is not significantly degraded from lack of document boundaries \cite{mueller2018pronountest,stojanovski-fraser-2019-improving} or random context \cite{voita2018anaphora}. To a large extent, both issues can be ignored, given the nature of our models. DomEmb is oblivious to the sentence order. CtxPool preserves some notion of sequentiality, but it should also be robust to these issues. Furthermore, we focus on obtaining domain signals. Even in an extreme case where the context comes from a different document (but from the same domain) we hypothesize similar performance. We later conduct an ablation study into whether arbitrary context from the same domain has a negative effect on performance. The results partially support our hypothesis by either matching or exceeding sentence-level performance, but also show that the correct context is important to obtain the best results.

\subsection{Baselines}

We compare our proposed methods against a sentence-level baseline (SentBase) and the domain tag (TagBase) approach \cite{kobus2017control}. We train TagBase with oracle domain tags, while at test time, we use tags obtained from a document-level domain classifier. All sentences within a document are marked with the same predicted domain tag. The domain classifier is a two-layer feed-forward network and the documents are represented as a bag-of-words. The classifier obtains an accuracy of 98.6\%. By design, documents from TED and PatTR were marked with tags from the remaining domains. Additionally, we compare with a context-aware model (CtxBase) which is similar to CtxPool, but we feed the full context to the context Transformer encoder, without applying max or average pooling beforehand. This model has token-level granular access to the context. We also train a concatenation model (ConcBase) \cite{tiedemann2017neural} using source-side context.

\section{Results}

\subsection{Zero-Resource Domain Adaptation}

In zero-resource domain adaptation experiments, we do not use any data from TED or PatTR, neither as training nor development data. The models are trained on our multi-domain dataset consisting of five domains. The results are shown in Table \ref{table:zero-resource-results}. We compute statistical significance with paired bootstrap resampling \cite{koehn2004statistical}.

SentBase achieves $16.7$ and $32.9$ BLEU on PatTR and TED respectively. The domains seen during training are more similar to TED in comparison to PatTR which is the reason for the large BLEU score differences. Our proposed models improve on PatTR by up to $0.4$ BLEU and on TED by up to $1.0$ BLEU. Improvements vary, but all models increase the BLEU score. The TagBase model does not improve significantly over SentBase.

\begin{table}[!ht]
	\normalsize
	\centering
	\def\arraystretch{1}
	\setlength\tabcolsep{0.6em}
	\begin{tabular}{ l | c | c}
		&  PatTR &  TED   \\ \hline
		SentBase     	 &  16.7  &  32.9  \\
		TagBase      	 &  16.8  &  33.0  \\
		DomEmb(max)  &  \phantom{*}\textbf{17.1}\dag  &  \phantom{*}\textbf{33.9}\dag  \\
		DomEmb(avg)  &  \phantom{*}\textbf{17.1}\dag  &  \phantom{*}33.8\dag  \\
		CtxPool(max) &  16.9  &  \phantom{*}33.6\ddag  \\
		CtxPool(avg) &  \phantom{*}\textbf{17.1}\dag  &  \phantom{*}\textbf{33.9}\dag  \\
	\end{tabular}
	\caption{Results on zero-resource domain adaptation for PatTR and TED. Best results in bold.
 	\dag - statistical significance with p $<$ 0.01, \ddag - p $<$ 0.05.}
	\label{table:zero-resource-results}
\end{table}

Our document-level models are robust across the two domains. These results confirm our assumption that access to document-level context provides for a domain signal. These models are oblivious to the actual characteristics of the domain since it was not seen in training, but presumably, they managed to match the zero-resource domain to a similar one. We assume that the reason for the larger improvements on TED in comparison to PatTR is that TED is a more similar domain to the domains seen  in training. As a result, matching TED to seen domains was easier for all models. Table \ref{table:zero-resource-results} shows that our proposed models improve on PatTR and TED and provides evidence that document-level context is useful for addressing zero-resource domains.

\begin{table*}
    \newcolumntype{Y}{>{\centering\arraybackslash}X}
	\normalsize
	\centering
	\def\arraystretch{0.92}
	\setlength\tabcolsep{0.35em}
	\begin{tabularx}{0.97\textwidth}{ l |  c c c c c c }
		domain & SentBase & TagBase & DomEmb(max) & DomEmb(avg) & CtxPool(max) & CtxPool(avg) \\ \hline
		\rule{0pt}{1em}Europarl       &  31.3 & 31.4 & \phantom{*}32.3\dag  & \phantom{*}\textbf{32.5}\dag & \phantom{*}32.4\dag & \phantom{*}32.3\dag  \\
		NewsComm        &  32.8 & 32.6 & 32.7 & 33.0 & \phantom{*}\textbf{33.1}\ddag & 32.8  \\
		OpenSub         &  26.6 & \phantom{*}27.1\ddag & \phantom{*}27.0\ddag  & \phantom{*}\textbf{27.5}\dag & \phantom{*}27.3\dag & \phantom{*}27.4\dag \\
		Rapid 		   &  40.7  & 40.9 & \phantom{*}41.1\ddag  & \phantom{*}41.5\dag & \phantom{*}41.4\dag & \phantom{*}\textbf{41.6}\dag  \\
		Ubuntu       &  31.5 & \phantom{*}\textbf{34.6}\dag & \phantom{*}32.8\ddag  & 31.9 & 31.6 & 32.1  \\
		\hline
		\rule{0pt}{1em}Average   & 30.4 & 30.9 & \textbf{31.0}  & \textbf{31.0} & 30.9 & \textbf{31.0}  \\
		Joint          &  29.1 & 29.2 & \phantom{*}29.5\dag  & \phantom{*}\textbf{29.8}\dag & \phantom{*}29.7\dag & \phantom{*}\textbf{29.8}\dag \\
	\end{tabularx}
	\caption{Results on the multi-domain dataset. Joint and average scores including PatTR and TED. Statistical significance computed for all scores except for Average. \dag - p $<$ 0.01, \ddag - p $<$ 0.05.}
	\label{table:multi-domain-results}
\end{table*}

\subsection{Evaluating Domains Seen During Training}

We assume that the improvements on zero-resource domains are because of document-level models having an increased capability to model domain. As a result, we also evaluate on the other domains which were seen during training. We show average BLEU and the BLEU score on the concatenation of all test sets. This is a useful way to evaluate in a multi-domain setting because it is less sensitive to larger improvements on a smaller test set.

Table \ref{table:multi-domain-results} shows the results. We first compare the baseline against DomEmb(avg). The smallest improvement is on NewsCommentary, only $0.2$ BLEU. Improvements vary between $0.8$ and $1.2$ BLEU on Europarl, OpenSubtitles and Rapid. On Ubuntu, this model improves only by $0.4$ BLEU. Joint and average BLEU improve by $0.7$ and $0.6$, respectively. Replacing average pooling with maximum pooling leads to slightly worse results on all domains except Ubuntu, but still improves upon the baseline. Our assumption is that averaging handles situations when there is a mix of domain signals because it can emphasize the more frequent domain signals. Max pooling is not able to differentiate between less and more frequent domain signals.

CtxPool(avg) and DomEmb(avg) perform similarly and have the same average and joint BLEU scores. Max pooling is slightly worse as shown by the performance of CtxPool(max). TagBase is not very effective in our experiments, improving slightly on some domains and only performing well on Ubuntu. We show that document-level context is useful for modeling multiple known domains at the same time. In the appendix we show translation examples from SentBase and DomEmb(avg).

\subsection{Context Length}

We also investigate the effect of context size on DomEmb(avg). Previous work on context-aware NMT \cite{zhang2018improving,miculicich2018documentnmt} typically showed that large context fails to provide for consistent gains. But this applies to more granular models which resemble the context-aware baseline CtxBase. In contrast, we observe that larger context does provide for improvements. We assume that for DomEmb, access to more context improves the likelihood of encountering domain-specific tokens.

\begin{table}[!ht]
	\normalsize
	\centering
	\def\arraystretch{1}
	\setlength\tabcolsep{0.3em}
	\begin{tabular}{ l | c | c | c}
		domain & \phantom{*}ctx=1\phantom{*} & \phantom{*}ctx=5\phantom{*} & ctx=10 \\ \hline
		Europarl         & 31.5\phantom{*}   & \phantom{\dag}32.0\dag\ddag   & \phantom{\ding{168}}\textbf{32.5}\dag\ding{168}   \\
		NewsComm   & 32.7\phantom{*}   & 32.9\phantom{*}   &  \textbf{33.0}\phantom{*}  \\
		OpenSub    & 26.8\phantom{*}   & \phantom{*}27.2*\ddag   &  \phantom{\ding{71}}\textbf{27.5}\dag\ding{71}  \\
		Rapid            & 41.1\ddag   & \phantom{*}\textbf{41.5}*\ddag   &  \textbf{41.5}*  \\
		Ubuntu 		     & 32.5\phantom{*}   & \phantom{*}\textbf{32.9}*\ddag   & 31.9\phantom{*}  \\
		\hline
		PatTR 		     & 17.0\ddag   & \textbf{17.2}\ddag   & 17.1\phantom{*}   \\
		TED	 	         & \phantom{*}33.5**   & 33.7\ddag   & \textbf{33.8}\phantom{*}  \\
		\hline
		\rule{0pt}{1em}Average 	   & 30.7\phantom{*}   &  \textbf{31.1}\phantom{*}  &  31.0\phantom{*}  \\
		Joint 		               & 29.3\ddag   &  \phantom{\ddag}29.7\dag\ddag  &  \phantom{\ding{168}}\textbf{29.8}\dag \ding{168}  \\
	\end{tabular}
	\caption{Results using the DomEmb(avg) model with different context sizes. Context size in number of previous sentences.
 	\ddag - p $<$ 0.01, ** - p $<$ 0.05, compared to SentBase. \dag - p $<$ 0.01, * - p $<$ 0.05, compared to \emph{ctx=1}. \ding{168} - p $<$ 0.01, \ding{71} - p $<$ 0.05, compared to \emph{ctx=5}.}
	\label{table:multi-domain-results-ctx-length}
\end{table}

We compare different context sizes and show the results in Table \ref{table:multi-domain-results-ctx-length}. A context size of 1 (\emph{ctx=1}) obtains the lowest scores on all domains. Using \emph{ctx=5} is comparable or slightly worse than \emph{ctx=10}. Both \emph{ctx=1} and \emph{ctx=5} get higher scores on Ubuntu and obtain significant improvements over SentBase on the full test set. Significance indicators for \emph{ctx=10} compared with respect to SentBase were already presented in Table \ref{table:multi-domain-results}. Due to resource limitations, we do not conduct a similar study for CtxPool.

\subsection{Comparison to Context-Aware Baselines}

Previous work on context-aware NMT has shown improvements in single-domain scenarios. In our work, we put two context-aware models to the test in a multi-domain setup. All models are trained with a 5 sentence context. The results in Table \ref{table:multi-domain-ctx-base} show that all models improve to varying degrees. They perform similarly on NewsCommentary and OpenSubtitles. CtxBase and ConcBase obtain better results on Europarl than DomEmb(avg) and worse on Ubuntu. CtxBase is best on Rapid. Both baselines obtained better scores on TED, showing they have some capacity to transfer to unseen domains. However, both failed to improve on PatTR.

\begin{table}[!ht]
	\normalsize
	\centering
	\def\arraystretch{1}
	\setlength\tabcolsep{0.4em}
	\begin{tabular}{ l | c | c | c}
		domain & CtxBase & ConcBase & DomEmb(a) \\ \hline
		Europarl         & \phantom{*}\textbf{32.4}\dag   & \phantom{*}\textbf{32.4}\dag
		& \phantom{*}32.0\dag \\
		NewsCo   & 32.8   & 32.7
		& \textbf{32.9} \\
		OpenSub    & \phantom{*}27.2\ddag   & \phantom{*}\textbf{27.4}\dag
        & \phantom{*}27.2\dag \\
		Rapid            & \phantom{*}\textbf{41.8}\dag   & 40.8
		& \phantom{*}41.5\dag \\
		Ubuntu 		     & 31.6   & 29.1
		& \phantom{*}\textbf{32.9}\dag \\
		\hline
		PatTR 		     & 16.6   & 14.8
		& \phantom{*}\textbf{17.2}\dag \\
		TED	 	         & \phantom{*}\textbf{34.1}\dag   & \phantom{*}\textbf{34.1}\dag
		& \phantom{*}33.7\dag \\
		\hline
		\rule{0pt}{1em}Average 	   & 30.9   &  30.2
		& \textbf{31.1} \\
		Joint 		               & \phantom{*}\textbf{29.7}\dag   &  29.5
		& \phantom{*}\textbf{29.7}\dag \\
	\end{tabular}
	\caption{Comparison with the context-aware baseline CtxBase and the concatenation model ConcBase.
 	\dag - p $<$ 0.01, \ddag - p $<$ 0.05 compared to SentBase.}
	\label{table:multi-domain-ctx-base}
\end{table}

We use 5 sentences of context for this experiment. Scaling the baseline models to large context is challenging with regards to computational efficiency and memory usage. In contrast, DomEmb scales easily to larger context. Furthermore, our analysis shows that DomEmb(avg) has the best average and joint score (CtxBase obtains the same joint score), improves on both unseen domains and consistently obtains significant improvements on all domains except NewsCommentary. As previous works show \cite{mueller2018pronountest}, these context-aware baselines improve fine-grained discourse phenomena such as anaphora resolution. We show in our manual analysis that DomEmb(avg) does not improve anaphoric pronoun translation which indicates that the improvements of our proposed model and the context-aware baselines are orthogonal.

\subsection{Translation of Domain-Specific Words}

We also evaluated the translation of domain-specific words. We extracted the most important words from a domain based on TF-IDF scores and selected the top 100 with the highest scores which have more than 3 characters. Next, we follow \newcite{liu2018homographs} and compute alignments using \emph{fastalign} \cite{dyer2013fast} based on the training set and force align the test set source sentences to the references and generated translations. We then compute the $F_1$ score of the translation of the domain-specific words. Results are shown in Table \ref{table:polysemy-results}. We compare SentBase with DomEmb(avg).

\begin{table}[!ht]
	\normalsize
	\centering
	\def\arraystretch{1}
	\setlength\tabcolsep{0.6em}
	\begin{tabular}{ l | c | c}
		&            SentBase & DomEmb(avg) \\ \hline
		Europarl    & 0.661 & \textbf{0.667} \\
		NewsComm    & 0.649 & \textbf{0.650} \\
		OpenSub     & 0.435 & \textbf{0.453} \\
		Rapid       & 0.724 & \textbf{0.730} \\
		Ubuntu      & 0.434 & \textbf{0.439} \\
		PatTR       & 0.407 & \textbf{0.409} \\
		TED         & 0.551 & \textbf{0.565} \\
	\end{tabular}
	\caption{$F_1$ score for domain-specific words.}
	\label{table:polysemy-results}
\end{table}

\begin{table}[!ht]
	\normalsize
	\centering
	\def\arraystretch{1}
	\setlength\tabcolsep{0.55em}
	\begin{tabular}{ l | c c}
		domain & SentBase & DomEmb(a)  \\ \hline
		PatTR	 	   & 34.4  & 34.4   \\
		\hline \hline
		& \multicolumn{2}{c}{ensemble} \\ \hline
		Europarl        & 29.0   & \phantom{*}\textbf{29.6}\dag  \\
		NewsCommentary  & 28.7   & \textbf{28.9}   \\
		OpenSubtitles   & 22.8   & \phantom{*}\textbf{23.4}\dag   \\
		Rapid 		    & 35.1   & \phantom{*}\textbf{35.7}\dag   \\
		Ubuntu          & 33.0   & \textbf{33.4}   \\
		PatTR 		    & 29.2   & \textbf{29.4}   \\
		TED 		    & 29.8   & \phantom{*}\textbf{30.4}\ddag   \\
		\hline
		\rule{0pt}{1em}Average        & 29.7  & \textbf{30.1}  \\
		Joint                         & 30.2  & \phantom{*}\textbf{30.6}\dag   \\
	\end{tabular}
	\caption{Domain adaptation results on PatTR for SentBase and DomEmb(avg). \dag - p $<$ 0.01, \ddag - p $<$ 0.05.}
	\label{table:domain-adaptation-results-patent}
\end{table}

\begin{table}[!ht]
	\normalsize
	\centering
	\def\arraystretch{1}
	\setlength\tabcolsep{0.55em}
	\begin{tabular}{ l | c c}
		domain & SentBase & DomEmb(a)  \\ \hline
		TED	              &  36.1  & \phantom{*}\textbf{36.6}\ddag     \\
		\hline \hline
		& \multicolumn{2}{c}{ensemble} \\ \hline
		Europarl          &  30.4  & \phantom{*}\textbf{30.8}\dag  \\
		NewsCommentary    &  31.9  & \phantom{*}\textbf{32.2}\ddag   \\
		OpenSubtitles     &  24.6  & \phantom{*}\textbf{25.4}\dag   \\
		Rapid 		      &  38.8  & \phantom{*}\textbf{39.5}\dag   \\
		Ubuntu            &  \textbf{32.7}  & 32.4   \\
		PatTR 		      &  16.9  & \phantom{*}\textbf{17.0}\ddag   \\
		TED 		      &  35.4  & \phantom{*}\textbf{35.8}\ddag   \\
		\hline
		\rule{0pt}{1em}Average        &  30.1 & \textbf{30.4}  \\
		Joint                         &  28.4 & \phantom{*}\textbf{28.8}\dag   \\
	\end{tabular}
	\caption{Domain adaptation results on TED for SentBase and DomEmb(avg). \dag - p $<$ 0.01, \ddag - p $<$ 0.05.}
	\label{table:domain-adaptation-results-ted}
\end{table}

\begin{table*}
	\normalsize
	\centering
	\def\arraystretch{0.92}
	\setlength\tabcolsep{0.5em}
	\begin{tabularx}{0.92\textwidth}{ l | c c c c c c c | c}
		domain & Europarl & NewsComm & OpenSub & Rapid & Ubuntu & PatTR & TED & True \\ \hline
		\rule{0pt}{1em}Europarl        & \textbf{31.3} & 30.1 & 30.6 & 30.3 & 30.7 & 30.7 & 30.7 & 32.5  \\
		NewsComm        & 30.6 & \textbf{32.8} & 31.9 & 30.1 & 32.3 & 31.5 & 32.1 & 33.0    \\
		OpenSub         & 22.2 & 23.1 & \textbf{27.1} & 22.0 & 25.4 & 24.4 & 26.7 & 27.5  \\
		Rapid 		    & 39.5 & 37.0 & 38.7 & \textbf{41.3} & 40.3 & 40.4 & 38.9 & 41.5   \\
		Ubuntu          & 29.3 & 29.1 & 29.2 & 29.6 & \textbf{31.4} & 31.1 & 30.1 & 31.9  \\
		PatTR           & 16.6 & 16.2 & 16.3 & 16.5 & 16.9 & \textbf{17.1} & 16.8 & 17.1  \\
		TED             & 30.0 & 33.0 & 33.1 & 28.8 & 33.4 & 31.5 & \textbf{33.7} & 33.8  \\
	\end{tabularx}
	\caption{Results from the ablation study investigating the influence of context from a different domain. Each row shows which domain is used as the test set and each column shows from which domain the context originates.}
	\label{table:ablation}
\end{table*}

DomEmb(avg) improved the $F_1$ score across all domains with the largest improvements on OpenSubtitles and TED. Our assumption is that the baseline translation of OpenSubtitles domain-specific words is more formal. A large part of the seen domains contain formal language in contrast to the informal subtitles. Lack of context seems to have biased SentBase to generate more formal translations. We later conduct a manual analysis on the TED test set where we confirm that word sense disambiguation is indeed improved in DomEmb(avg).

\subsection{Domain Adaptation with Available In-Domain Data}

We also conduct a classical domain adaptation evaluation where access to in-domain data is allowed. We either use PatTR or TED as in-domain data and evaluate with SentBase and DomEmb(avg). In both cases we consider the concatenation of the remaining domains as out-of-domain. This setup differs from zero-resource domain adaptation because we assume access to in-domain training and dev data.

First, we train the baseline and DomEmb(avg) on out-of-domain data. Since these initial models are identical to the ones in the zero-resource setup, we reuse them. We then continue training on the corresponding in-domain data. Table \ref{table:domain-adaptation-results-patent} shows the results for PatTR. Fine-tuning the baseline and DomEmb(avg) on PatTR improves BLEU by a large margin, both obtaining $34.4$ BLEU. The results are unsurprising because our model is tailored to multi-domain setups and is unlikely to contribute to large improvements when fine-tuning on a single domain. Identifying the domain in such a case is trivial and using large context should not be helpful.

The strengths of our approach come to light by comparing it against SentBase in an ensembling scenario as in \newcite{freitag2016fast}. We ensemble DomEmb(avg) trained on out-of-domain data with DomEmb(avg) fine-tuned on in-domain data and do the same for SentBase. The DomEmb(avg) ensemble is better than the SentBase ensemble on all domains and on joint BLEU. Similar results are obtained when fine-tuning on TED which are shown in Table \ref{table:domain-adaptation-results-ted}. 

\subsection{Ablation}

We previously hypothesized that our models will benefit from context from different documents within the same domain. We conduct an ablation study to test this assumption using DomEmb(avg) model, similar to the study in \cite{kobus2017control}, where they investigated the effect of giving the wrong domain tag to every sentence.

For DomEmb(avg), we simulate this approach by replacing the real contextual representation of each test sentence with $C_d$, which is context representative of domain $d$. We first compute $C_d^{'} = \frac{1}{N_d} \sum_{i=1}^{N_d} c_i^d$ where $c_i^d$ is the contextual representation of a test sentence in domain $d$ and $N_d$ is the number of test sentences in $d$. $c_i^d$ is the average of the context token-level embeddings for sentence $i$. Finally, $C_d = \argmax_{c_i^d} cos(c_i^d, C_d^{'})$. This procedure is conducted for each domain $d$ separately.  

Table \ref{table:ablation} shows the results. On OpenSubtitles, Rapid, PatTR and TED, DomEmb(avg) improves on the sentence-level baseline if presented with context from the same domain (which is usually not from the same document). On Europarl, NewsCommentary and Ubuntu, it performs similarly to the baseline. In almost all cases, providing a mismatched context degrades the performance of the original DomEmb(avg). The results show that the model is relatively robust to incorrect but closely related context which provides evidence for our hypothesis that DomEmb captures domain-relevant features. However, the correct context is important to obtain the best results across all domains. 
Our finding is in contrast with recent results \cite{li2020multi-encoder} where they show that multi-encoder context-aware NMT models do not encode contextual information.

\subsection{Manual Analysis}

We conduct a manual analysis of SentBase and DomEmb(avg) by inspecting them on the TED test set. We only consider translation differences related to word senses and ignore other types of mistakes. We find 156 cases where the two models translate a word in a different sense and at least one of them outputs the correct sense. We define 3 categories: (i) one model is correct while the other wrong; (ii) both are correct, but one is closer to the actual meaning and (iii) both are correct, but one matches the reference translation. DomEmb(avg) is better on (i) in 43 cases as opposed to the 19 cases where SentBase is better. The ratio of DomEmb(avg) being correct in contrast to SentBase is 23/12 in (ii) and 38/21 in (iii). This shows that DomEmb(avg) is better at coherence which is closely related to better domain modeling in multi-domain setups where the number of probable senses is larger than in a single domain. Furthermore, we find that DomEmb(avg) does not improve on pronoun translation. In fact, in several cases it introduced errors, thus ruling out better coreference resolution as a source of improvements. 

\section{Conclusion}

We presented document-level context-aware NMT models and showed their effectiveness in addressing zero-resource domains. We compared against strong baselines and showed that document-level context can be leveraged to obtain domain signals. The proposed models benefit from large context and also obtain strong performance in multi-domain scenarios. Our experimental results show the proposed models obtain improvements of up to $1.0$ BLEU in this difficult zero-resource domain setup. Furthermore, they show that document-level context should be further explored in future work on domain adaptation and  suggest that larger context would be beneficial for other discourse phenomena such as coherence.

\section*{Acknowledgments}

This work was supported by the European Research Council (ERC) under the
European Union’s Horizon 2020 research and innovation programme (grant
agreement No.~640550) and by the German Research Foundation (DFG; grant FR
2829/4-1).

\bibliography{eacl2021}
\bibliographystyle{acl_natbib}

\clearpage
\appendix

\section{Preprocessing and Hyperparameters}

We tokenize all sentences using the script from Moses\footnote{\url{https://github.com/moses-smt/mosesdecoder/blob/master/scripts/tokenizer/tokenizer.perl}}. We apply BPE splitting\footnote{\url{https://github.com/rsennrich/subword-nmt}} with 32K merge operations. We exclude TED and PatTR when computing the BPEs. The BPEs are computed jointly on the source and target data. Samples where the source or target are larger than 100 tokens are removed. We also apply a per-sentence limit of 100 tokens on the context, meaning that models trained on 10 sentences of context have a limit of 1000 tokens. A batch size of 4096 is used for all models.

We first train a sentence-level baseline until convergence based on early-stopping. All context-aware models are initialized with the parameters from this pretrained sentence-level baseline. Parameters that are specific to the models' architectures are randomly initialized. All proposed models in this work share the source, target, output and context embeddings. The models' architecture is a 6 layer encoder/decoder Transformer with 8 attention heads. The embedding and model size is 512 and the size of the feed-forward layers is 2048. The number of parameters for all models is shown in Table \ref{table:num-parameters}. We use label smoothing with 0.1 and dropout in the Transformer of 0.1. Models are trained on 2 GTX 1080 Ti GPUs with 11GB RAM.

\begin{table}[ht!]
	\normalsize
	\centering
	\def\arraystretch{1}
	\setlength\tabcolsep{0.6em}
	\begin{tabular}{ l | c}
		Model         & parameters \\ \hline
		SentBase & 61M \\
		CtxBase  & 74M \\
		CtxPool    & 74M \\
		DomEmb(avg) & 63M \\
	\end{tabular}
	\caption{Number of model parameters. TagBase, ConcBase and DomEmb(max) have the same number of parameters as SentBase.}
	\label{table:num-parameters}
\end{table}

The initial learning rate for the document-level models is $10^{-4}$. For the classical domain adaptation scenario with fine-tuning, we use a learning rate of $10^{-5}$ in order not to deviate too much from the well-initialized out-of-domain model. We lower the learning rate by a factor of 0.7 if no improvements are observed on the validation perplexity in 8 checkpoints. A checkpoint is saved every 4000 updates. We did not do any systematic hyperparameter search.

Before inference, we average the parameters of the 8 best checkpoints based on the validation perplexity. We use a beam size of 12. BLEU scores are computed on detokenized text using \emph{multi-bleu-detok.perl} from the Moses scripts\footnote{\url{https://github.com/moses-smt/mosesdecoder/blob/master/scripts/generic/multi-bleu-detok.perl}}. For the evaluation of translation of domain-specific words, we used the script from \cite{liu2018homographs}\footnote{\url{https://github.com/frederick0329/Evaluate-Word-Level-Translation}}.

\begin{table*}
    \newcolumntype{Y}{>{\centering\arraybackslash}X}
	\normalsize
	\centering
	\def\arraystretch{0.92}
	\setlength\tabcolsep{0.35em}
	\begin{tabularx}{0.97\textwidth}{ l |  c c c c c c }

		domain & SentBase & TagBase & DomEmb(max) & DomEmb(avg) & CtxPool(max) & CtxPool(avg) \\ \hline
		\rule{0pt}{1em}Europarl       &  33.3 & 33.6 & 33.6  & 33.7 & 33.8 & 33.8  \\
		NewsComm        &  34.1 &	34.3 &	34.1 &	34.1 &	34.2 &	34.1  \\
		OpenSub         &  33.3 &	34.2 &	34.2 &	34.5 &	34.1 &	34.2 \\
		Rapid 		   &  39.4 &	39.7 &	39.5 &	39.7 &	39.8 &	39.9  \\
		Ubuntu       &  40.2 &	43.0 &	41.3 &	42.6 &	42.0 &	42.2 \\

	\end{tabularx}
	\caption{BLEU scores on the development sets of the multi-domain dataset.}
	\label{table:multi-domain-results-dev}
\end{table*}

\section{Datasets}

We use the document-aligned versions of Europarl, NewsCommentary and Rapid from WMT 2019\footnote{\url{http://statmt.org/wmt19/translation-task.html}}. We also use OpenSubtitles\footnote{\url{http://opus.nlpl.eu/OpenSubtitles-v2018.php}}\footnote{\url{http://www.opensubtitles.org/}} \cite{lison2016opensubtitles2016}, Ubuntu\footnote{\url{http://opus.nlpl.eu/Ubuntu.php}}, PatTR\footnote{\url{http://www.cl.uni-heidelberg.de/statnlpgroup/pattr/}} and TED\footnote{\url{https://wit3.fbk.eu/2015-01}}.

\begin{table}[ht!]
	\normalsize
	\centering
	\def\arraystretch{1}
	\setlength\tabcolsep{0.3em}
	\begin{tabular}{ l | c | c | c}

		domain & ctx=1 & ctx=5 & ctx=10 \\ \hline
		Europarl         & 33.5	& 33.8 & 33.7   \\
		NewsComm   & 34.0 &	34.2 & 34.1  \\
		OpenSub    & 33.7 &	34.1 & 34.5  \\
		Rapid            & 39.7 &	39.8 & 39.7 \\
		Ubuntu 		     & 41.5 &	43.0 & 42.6 \\

		\hline \hline
		domain & CtxBase & ConcBase & DomEmb(a) \\ \hline
		Europarl         & 34.0	& 34.1 & 33.7   \\
		NewsComm   & 34.0 &	33.9 & 34.1  \\
		OpenSub    & 33.9 & 34.5 & 34.5  \\
		Rapid            & 40.1 & 39.1 & 39.7 \\
		Ubuntu 		     & 42.3 & 42.3 & 42.6 \\

	\end{tabular}
	\caption{Results on the development sets using the DomEmb(avg) model with different context sizes and comparing DomEmb(avg) with \emph{ctx=10} against CtxBase and ConcBase.}
	\label{table:multi-domain-results-ctx-length-dev}
\end{table}

\begin{table}[ht!]
	\normalsize
	\centering
	\def\arraystretch{1}
	\setlength\tabcolsep{0.55em}
	\begin{tabular}{ l | c c}

		domain & SentBase & DomEmb(a)  \\ \hline
		TED        & 33.2 &	33.4  \\ \hline
		PatTR	 	   & 36.4 & 36.3   \\
	\end{tabular}
	\caption{Domain adaptation results on PatTR and TED for SentBase and DomEmb(avg) on the development sets.}
	\label{table:domain-adaptation-results-dev}
\end{table}

\section{Validation performance}

In Table \ref{table:multi-domain-results-dev}, Table \ref{table:multi-domain-results-ctx-length-dev} and Table \ref{table:domain-adaptation-results-dev} we present BLEU scores on the development sets for all the experiments we ran. We only show results for the sets we actually used during training and therefore ignore TED and PatTR for which we had no access to data at training time. The results for TagBase are with oracle domain tags. For the experiments with continued training on TED and PatTR, we show results only on the development sets for TED and PatTR.

\section{Computational Efficiency}

In this section, we compare the computational efficiency of our proposed methods. We compare how many seconds on average are needed to translate a sentence from the test set. The average times are $0.2588$, $0.2763\pm0.0124$, $0.3662$ for SentBase, DomEmb and CtxPool, respectively. DomEmb is insignificantly slower than the sentence-level baseline, in contrast to CtxPool, which is to be expected considering the additional applying of self-attention over the compressed context. In terms of training time, SentBase converged after 90 hours of training, DomEmb(avg) after 168h and CtxPool(avg) after 116h.

\section{Examples}

In Table \ref{table:manual-examples} we show some example translations from the sentence-level baseline and our DomEmb(avg) model. We show examples where our model corrected erroneous translations from the baseline. Some of the proper translations should be evident from the main sentence itself, but some can only be inferred from context. The first four examples are from TED and the last from PatTR.

In the first example, we can see that the sentence-level baseline translates ``students'' as ``Studenten'' (university students), but the correct translation in this case is ``Sch{\"u}ler'' (elementary or high school student). The main sentence itself is not informative enough for the sentence-level model to make this distinction. In contrast, the DomEmb model has access to more information which provides for the appropriate bias towards the correct translation.

The second sentence depicts an example where it's nearly impossible for the baseline to make a correct prediction for the translation of ``ambassador'' because it depends on whether the person is male (Botschafter) or female (Botschafterin). In the third example, the sentence-level model translated ``model'' as in ``a role model'' (Vorbild), but the context indicates that the speaker talks about ``fashion models''.

Examples 4 and 5 are relatively unintuitive because the main sentences themselves should be enough to infer the correct translation. In example 4, ``reflect'' refers to the physical process of reflection and should not be translated as in ``to reflect on oneself'' (``denken''), while in example 5, ``raise'' refers to the action of ``lifting'' or ``elevating''(``aufw{\"a}rtsbewegt'' or ``hochzuziehen'') some object instead of ``raising'' as in ``raising a plant (from a seed)'' (``z{\"u}chten''). 

The last example shows that the sentence-level model translates ``springs'' (``Federn'' which is a part of the compound word ``Druckfedern'' in the reference) as in ``water springs'' (``Quellen'' which is a part of the compound word ``Kompressionsquellen'') while it should be translated instead as in the physical elastic device. However, in other test sentences, both SentBase and DomEmb(avg) translated ``spring'' as a season, even though this should be less likely in PatTR, showing that our model does not always succeed in capturing domain perfectly.

\begin{table*}
	\small
	\centering
	\def\arraystretch{1.28}
	\setlength\tabcolsep{0.6em}

	\begin{tabular}{| p{15cm} |}
		\hline

		\textit{Source} \\
		We all knew we were risking our lives -- the teacher, the \textbf{students} and our parents. \\
		\textit{Reference} \\
		Wir alle wussten, dass wir unser Leben riskierten: Lehrer, \textbf{Sch{\"u}ler} und unsere Eltern. \\
		\textit{SentBase} \\
		Wir alle wussten, dass wir unser Leben riskieren... den Lehrer, die \textbf{Studenten} und unsere Eltern. \\
		\textit{DomEmb(avg)} \\
		Wir wussten alle, dass wir unser Leben riskierten. Der Lehrer, die \textbf{Sch{\"u}ler} und unsere Eltern. \\
		\hline

		\textit{Source} \\
		That\textquotesingle s why I am a global \textbf{ambassador} for 10x10, a global campaign to educate women.\\
		\textit{Reference} \\
		Deshalb bin ich globale \textbf{Botschafterin} f{\"u}r 10x10, einer weltweiten Kampagne f{\"u}r die Bildung von Frauen. \\
		\textit{SentBase} \\
		Aus diesem Grund bin ich ein globaler \textbf{Botschafter} f{\"u}r 10x10, eine weltweite Kampagne zur Ausbildung von Frauen. \\
		\textit{DomEmb(avg)} \\
		Deshalb bin ich eine globale \textbf{Botschafterin} f{\"u}r 10x10, eine weltweite Kampagne zur Ausbildung von Frauen. \\
		\hline

		\textit{Source} \\
		And I am on this stage because I am a \textbf{model}.\\
		\textit{Reference} \\
		Und ich stehe auf dieser B{\"u}hne, weil ich ein \textbf{Model} bin. \\
		\textit{SentBase} \\
		Und ich bin auf dieser B{\"u}hne, weil ich ein \textbf{Vorbild} bin.\\
		\textit{DomEmb(avg)} \\
		Und ich bin auf dieser B{\"u}hne, weil ich ein \textbf{Model} bin.\\
		\hline

		\textit{Source} \\
		It\textquotesingle s going to bounce, go inside the room, some of that is going to \textbf{reflect} back on the door ...\\
		\textit{Reference} \\
		Es wird abprallen, in den Raum gehen, ein Teil davon wird wieder zur{\"u}ck auf die T{\"u}r \textbf{reflektiert} ... \\
		\textit{SentBase} \\
		Es wird abprallen, ins Zimmer gehen, etwas davon wird wieder an die T{\"u}r \textbf{denken} ...\\
		\textit{DomEmb(avg)} \\
		Es wird abprallen, ins Zimmer gehen, etwas davon wird wieder {\"u}ber die T{\"u}r \textbf{reflektieren} ...\\
		\hline

		\textit{Source} \\
		Tie member 60 is driven to \textbf{raise} movable cone 58 ...\\
		\textit{Reference} \\
		Mit dem Zugelement 60 wird durch den An der bewegliche Kegel 58 \textbf{aufw{\"a}rtsbewegt} ... \\
		\textit{SentBase} \\
		Tie-Mitglied 60 wird angetrieben, bewegliche Konfit{\"u}re 58 zu \textbf{z{\"u}chten} ...\\
		\textit{DomEmb(avg)} \\
		Teemitglied 60 wird angetrieben, bewegliche Kegel 58 \textbf{hochzuziehen} ...\\
		\hline

		\textit{Source} \\
		It is only when a certain pressure level is reached that the pistons are pushed back against the action of the \textbf{compression springs} ... \\
		\textit{Reference} \\
		Erst bei Erreichen eines bestimmten Druckniveaus werden die Kolben gegen die Wirkung der \textbf{Druckfedern} zur{\"u}ckgeschoben ... \\
		\textit{SentBase} \\
		Erst wenn ein gewisses Druckniveau erreicht ist, werden die Pistonen gegen die Wirkung der \textbf{Kompressionsquellen} zur{\"u}ckgedr{\"a}ngt ...\\
		\textit{DomEmb(avg)} \\
		Erst wenn ein bestimmtes Druckniveau erreicht ist, werden die Pistonen gegen die Wirkung der \textbf{Kompressionsfedern} zur{\"u}ckgedr{\"a}ngt ...\\
		\hline

	\end{tabular}
	\caption{Example translations obtained using sentence-level baseline and the DomEmb(avg) model. Relevant parts of the examples are in bold.}
	\label{table:manual-examples}
\end{table*}

\end{document}